\documentclass{article} 
\usepackage{iclr2027_conference,times}

\usepackage{amsmath,amsfonts,bm}

\def\eqref#1{equation~\ref{#1}}

\def\1{\bm{1}}

\DeclareMathAlphabet{\mathsfit}{\encodingdefault}{\sfdefault}{m}{sl}
\SetMathAlphabet{\mathsfit}{bold}{\encodingdefault}{\sfdefault}{bx}{n}

\usepackage{hyperref}
\usepackage{url}
\usepackage{graphicx}
\usepackage{booktabs}
\usepackage{longtable}
\usepackage{arydshln} 

\title{INCONVENIENT: A Benchmark to Assess LLM Generalization Capabilities from Convenience Samples}

\author{\mdseries
\begin{tabular}[t]{@{}l@{\hspace{6em}}l@{}}
\textbf{Sara Rezaeimanesh}   & \textbf{Chad Klochko, M.D.} \\
Michigan State University    & Henry Ford Health \\
East Lansing, MI, USA        & Detroit, MI, USA \\
\texttt{rezaeima@msu.edu}    & \texttt{chadk@rad.hfh.edu} \\[3ex]
\textbf{Kundan Thind, Ph.D.} & \textbf{Mohammad Ghassemi} \\
Henry Ford Health            & Michigan State University \\
Detroit, MI, USA             & East Lansing, MI, USA \\
\texttt{kthind1@hfhs.org}    & \texttt{ghassem3@msu.edu}
\end{tabular}
}

\newcommand{\name}{INCONVENIENT}

\iclrfinalcopy 
\begin{document}

\maketitle

\begin{abstract}
Conventional machine learning models for tabular prediction typically assume that training data are representative of the population on which they will be deployed. In practice, however, data are often collected based on convenience, availability, accessibility, or cost, resulting in systematic sampling biases that can distort feature--target relationships and undermine real-world generalization. Existing tabular distribution-shift benchmarks primarily construct out-of-distribution (OOD) settings by partitioning observations across naturally occurring domains. While such partitions may induce changes in \(P(Y \mid X)\), the resulting shifts are neither explicitly controlled nor isolated, and the evaluation distributions themselves are often not representative of the broader target population. We introduce \name, a benchmark designed to study robustness to convenience-sample bias. Rather than relying on naturally occurring domain partitions, \name\ directly perturbs feature--target relationships in the training distribution while preserving the original population distribution as the deployment environment. This setting more closely reflects real-world scenarios in which models are developed from unrepresentative samples and subsequently deployed on a broader population. The distinction is particularly relevant for large language models (LLMs), whose pretraining may encode population-level knowledge that extends beyond the relationships observed in a biased training dataset. Unlike existing distribution-shift benchmarks, \name\ provides explicit control over both the direction and severity of relationship shifts, enabling a systematic assessment of robustness to convenience sampling. The benchmark comprises 200 tabular classification and regression datasets spanning diverse application domains and generates around $\sim15{,}500$ shifted training distributions which lead to substantial downstream performance degradation.

\end{abstract}

\section{Introduction}

Machine learning models are typically developed under the assumption that training data provide a representative sample of the population on which the model will ultimately be deployed. In practice, however, this assumption is rarely satisfied. Data are often collected based on convenience, availability, accessibility, or cost rather than through representative sampling procedures. Medical datasets may be collected from a single hospital system, consumer datasets may overrepresent users of a particular platform, and scientific datasets may be limited to populations that are easiest to recruit. As a result, the relationships observed in the training data can differ from those that hold in the broader population.

LLMs are pretrained on massive corpora that may encode broad statistical knowledge about real-world relationships. When presented with a biased or unrepresentative dataset, an LLM-based prediction system may possess prior knowledge that conflicts with the relationships observed in the training sample. 

A large body of work has studied the impact of distribution shift on machine learning systems. Benchmarks such as WILDS \citep{koh2021wilds} demonstrated that modern models often struggle to generalize across naturally occurring
shifts in vision, language, and graph domains. In the tabular setting,
TableShift \citep{gardner2023tableshift} and WhyShift \citep{liu2023whyshift}
construct out-of-distribution evaluation settings by partitioning data into
distinct real-world domains, such as geographic regions, institutions, or time
periods. These benchmarks therefore evaluate generalization across naturally
occurring population differences that induce complex combinations of covariate,
label, and conditional shifts; WhyShift in particular decomposes each observed
shift into its covariate and concept components and finds that concept shift
dominates on many real tabular tasks. However, in these benchmarks, both the training and evaluation distributions change as a consequence of the shift. As a result, poor performance may reflect difficulties adapting to a new target population rather than an inability to recover population-level relationships obscured by biased sampling. While these settings capture important real-world deployment challenges, they leave a common form of dataset bias largely unaddressed: scenarios in which the deployment population remains fixed, but the training sample is systematically unrepresentative of that population. More recently, TabFSBench \citep{cheng2025tabfsbench} introduced a controlled benchmark that synthetically generates feature shifts through perturbations of the observed feature space. However, it does not explicitly isolate shifts in the underlying feature--target relationship, which is the primary focus of our benchmark.

To address this limitation, we introduce \textbf{\name}, a benchmark for evaluating robustness to convenience-sample bias. Rather than constructing train and test sets from different domains, \name\ begins with a representative population dataset and systematically constructs unrepresentative training samples that distort the relationship between features and targets. Models are then evaluated on test examples drawn from the original population. \name\ generates controlled shifts by manipulating feature--target relationships in the training data through coefficient amplification, attenuation, and sign-reversal operations. By explicitly controlling both the direction and severity of these perturbations, \name\ enables the systematic study of robustness to sampling-induced relationship shifts. It also provides control over severity of shift which allows comprehensive analysis of how sever of a shift and what kind of shift is a model more robust to handle. The benchmark contains 200 tabular prediction datasets spanning diverse application domains and yields around 15{,}500 distribution-shift scenarios. Across these tasks, convenience-sample training distributions induce substantial degradation in downstream predictive performance, creating a challenging testbed for evaluating robustness under sampling-induced shift.

\section{Related Work}

\paragraph{Sample selection bias and convenience sampling.}
The observation that non-random sampling distorts learned relationships predates machine learning. \citet{heckman1979sample} formalized selection bias in econometrics, and the learning-theoretic treatment of \citet{zadrozny2004learning} and \citet{cortes2008sample} showed that selection depending on $X$ alone (the ``missing at random'' or covariate-shift case) is correctable by importance weighting, whereas selection that depends on $Y$ or on the $X$--$Y$ relationship is not identifiable from the biased sample. 

\paragraph{Distribution shift benchmarks.}
WILDS \citep{koh2021wilds} established the now-standard recipe for evaluating OOD generalization: partition a dataset along a naturally occurring axis such as hospital, country, or time, train on some partitions, and test on others. Tabular benchmarks have largely inherited this design. TableShift \citep{gardner2023tableshift} assembles tasks with domain-based splits and finds that the ID--OOD gap tracks label shift, while WhyShift \citep{liu2023whyshift} goes further and decomposes each observed shift into its covariate ($Y|X$-preserving) and concept ($Y|X$-changing) components, finding that concept shift dominates on many real tabular tasks. TabFSBench \citep{cheng2025tabfsbench} takes a complementary, synthetic approach and removes or adds feature columns between training and test, but it does not target the feature--target relationship itself. \name\ differs from both families on two axes. First, the deployment distribution is held fixed at the population, so degradation is attributable to the training sample alone. Second, the shift is constructed by directly manipulating the coefficient of a chosen feature so its direction and severity are known and controllable.

\paragraph{LLMs and pretrained priors for tabular prediction.}
A growing body of work suggests that large language models (LLMs) encode knowledge that can improve learning under distribution shift and limited data. TabLLM \citep{hegselmann2023tabllm} and LIFT \citep{dinh2022lift} serialize tabular instances as text and demonstrate that LLMs can outperform traditional machine learning models in low-data regimes. A complementary line of research seeks to extract this knowledge as an explicit prior rather than using the LLM directly as a predictor. LLM Processes \citep{requeima2024llm} condition predictive distributions on natural-language descriptions of the task, AutoElicit \citep{capstick2025autoelicit} elicits priors over linear-model coefficients from feature names alone, and LoID \citep{rezaeimanesh2026languagemodelsknowdont} extracts feature-wise Bayesian priors directly from model logits without relying on generated text. Across a variety of tabular prediction tasks, these approaches often match or exceed expert-derived priors and improve performance under data scarcity. Complementing these findings, \citet{selby2024quantitative} evaluate the extent to which elicited quantities align with real-world ground truth relationships.

This literature provides evidence that LLMs encode population-level knowledge about relationships between semantically meaningful features and targets. However, existing evaluations largely focus on settings where the training data are limited, and do not address whether such knowledge remains useful when the observed data actively contradict those relationships. \name\ is designed to address this gap. By constructing training samples in which feature--target relationships are systematically weakened, amplified, or even reversed while preserving the original population as the evaluation distribution, the benchmark isolates scenarios where recovering the correct population relationship requires leveraging prior knowledge beyond the observed sample.

\section{Benchmark Creation}

\subsection{Problem Statement}

Consider a tabular dataset composed of examples $(x, y, s) \sim P_s$ where $x \in \mathbb{R}^d$ is the input feature vector, $y$ is the prediction target (either categorical for classification or continuous for regression), and $s \in \{0, 1\}$ is the sampling indicator denoting whether the example is drawn from a random sample ($s=1$) or a convenience sample ($s=0$). 

We can view the overall data distribution as a mixture of two sampling distributions: the population distribution $P_{\text{pop}}$ representing random samples from the target population, and the convenience distribution $P_{\text{conv}}$ representing systematically biased samples. The test distribution is always drawn from the population: $P_{\text{test}} = P_{\text{pop}}$.

In the random-sampling setting, our goal is to learn a predictor $f_\theta: \mathbb{R}^d \to \mathcal{Y}$ that accurately predicts $y$ using examples drawn from $P_{\text{pop}}$. This represents the ideal scenario where training data is representative of the deployment population.

A distribution shift occurs because $P_{\text{conv}} \neq P_{\text{pop}}$. As a consequence of this shift, the joint distributions $p_{\text{conv}}(x, y) \neq p_{\text{pop}}(x, y)$ differ between convenience and population samples. For a fixed predictor $f_\theta$ trained on a convenience sample, we define the \textit{convenience sample gap} as:
\begin{equation}
\Delta_{\text{perf}} = \text{Perf}(f_\theta^{\text{pop}}, P_{\text{test}}) - \text{Perf}(f_\theta^{\text{conv}}, P_{\text{test}})
\label{eq:shift_gap}
\end{equation}
where $f_\theta^{\text{pop}}$ is trained on random samples from $P_{\text{pop}}$, $f_\theta^{\text{conv}}$ is trained on convenience samples from $P_{\text{conv}}$, and $\text{Perf}(\cdot)$ denotes the performance metric (AUC-ROC for classification, R² for regression). Both models are evaluated on held-out test data from $P_{\text{test}}$. The gap $\Delta_{\text{perf}}$ quantifies the performance degradation due to training on convenience samples rather than random samples. 

In our setting, we assume that no information about the specific form of distribution shift is available during training – i.e., there is no knowledge of how $p(y)$, $p(y|x)$, or $p(x)$ differ between $P_{\text{conv}}$ and $P_{\text{pop}}$, and no unlabeled data from the population distribution is accessible. This reflects the realistic scenario where practitioners train models on available convenience samples without knowing the precise nature of sampling biases.

\subsection{Dataset selection}
Because the primary goal of \name\ is to evaluate methods that leverage prior knowledge encoded in large language models, we restrict our benchmark to datasets with semantically meaningful feature representations. Consequently, we exclude image and audio datasets where individual input dimensions typically lack interpretable semantic meaning and therefore provide limited opportunity for methods based on feature names, metadata, or task descriptions to exploit external knowledge. 

Our benchmark comprises 200 tabular datasets spanning a diverse range of application domains. The datasets come from OpenML \citep{vanschoren2014openml}, Kaggle, and TALENT \citep{liu2024talent}. The collection includes 80 regression datasets, 39 binary classification datasets, and 81 multiclass classification datasets. Dataset sizes range from hundreds to tens of thousands of observations, and the number of features varies from fewer than 10 to more than 100. This diversity enables the evaluation of robustness to convenience-sample bias across a broad spectrum of prediction tasks and data regimes.

\subsubsection{Reference model}

For each dataset, we first split the data into a training pool (50\%) and a held-out test set (50\%). We fit a reference linear model on the entire training pool to obtain baseline coefficients $\beta_{\text{ref}} = (\beta_1, \ldots, \beta_d)$, where $d$ is the number of features after preprocessing (including one-hot encoded categorical variables). For classification tasks, we use logistic regression with $C=\infty$ (no regularization), yielding coefficients in the log-odds space. For regression tasks, we use ordinary least squares linear regression.

\subsubsection{Coefficient manipulation operations}

We define three types of coefficient shifts, each creating a distinct form of distribution shift:

\paragraph{Amplification.} Increase the magnitude of target feature coefficients by a multiplicative factor $\alpha > 1$. The target criterion is:
\begin{equation}
\beta_{\text{candidate}, j} > \alpha \cdot \beta_{\text{ref}, j}, \quad \forall j \in \mathcal{F}_{\text{target}}
\end{equation}
The influence weight is $w_j(\text{amplify}) = \text{sign}(\beta_{\text{ref}, j})$, prioritizing examples that strengthen the existing coefficient direction. We test amplification factors $\alpha \in \{1.15, 1.3, 1.5, 2.0, 3.0, 5.0\}$, creating convenience samples where feature-target relationships are exaggerated compared to the population.

\paragraph{Attenuation.} Decrease the magnitude of target feature coefficients by a multiplicative factor $0 < \gamma < 1$. The target criterion is:
\begin{equation}
|\beta_{\text{candidate}, j}| < \gamma \cdot |\beta_{\text{ref}, j}|, \quad \forall j \in \mathcal{F}_{\text{target}}
\end{equation}
The influence weight is $w_j(\text{attenuate}) = -\text{sign}(\beta_{\text{ref}, j}) \cdot \text{sign}(x_{ij})$, prioritizing examples that weaken coefficient magnitudes. We test attenuation factors $\gamma \in \{0.2, 0.4, 0.6, 0.7, 0.9\}$, creating convenience samples where feature-target relationships are diluted or noisier than in the population.

\paragraph{Sign flip.} Reverse the direction of target feature effects by flipping coefficient signs. The target criterion is:
\begin{equation}
\text{sign}(\beta_{\text{candidate}, j}) \neq \text{sign}(\beta_{\text{ref}, j}), \quad \forall j \in \mathcal{F}_{\text{target}}
\end{equation}
The influence weight is $w_j(\text{flip}) = -\text{sign}(\beta_{\text{ref}, j})$, prioritizing examples that induce sign reversals. This creates extreme distribution shifts, where convenience samples exhibit reversed relationships compared to the population (e.g., marijuana use in the head-and-neck cancer data of \citet{rezaeimanesh2026languagemodelsknowdont}).

\subsubsection{Greedy subset selection algorithm}

To create convenience training sets with specific coefficient shifts, we employ a greedy iterative algorithm that selects training examples to achieve a target coefficient configuration. The algorithm operates as follows:

\begin{enumerate}
    \item \textbf{Influence scoring}: For each example $(x_i, y_i)$ in the training pool, compute an influence score $s_i$ measuring how much including that example would push the target features' coefficients in the desired direction. For classification, the influence is based on the gradient contribution weighted by prediction uncertainty:
    \begin{equation}
    s_i = \sum_{j \in \mathcal{F}_{\text{target}}} (y_i - \hat{p}_i) \cdot x_{ij} \cdot w_j(\text{direction})
    \end{equation}
    where $\hat{p}_i$ is the predicted probability under the reference model, and $w_j(\text{direction})$ encodes the desired direction of change. For regression, the influence uses residuals: $s_i = \sum_{j \in \mathcal{F}_{\text{target}}} (y_i - \hat{y}_i) \cdot x_{ij} \cdot w_j(\text{direction})$.

    \item \textbf{Initial candidate set}: Sort all training pool examples by influence score in descending order. Initialize the candidate subset with the top $n_{\text{min}} = \lfloor \alpha_{\text{min}} \cdot N_{\text{pool}} \rfloor$ highest-influence examples, where $\alpha_{\text{min}}$ is the minimum fraction hyperparameter and $N_{\text{pool}}$ is the training pool size.

    \item \textbf{Iterative expansion}: Fit a logistic regression (or linear regression) model on the current candidate subset and compute its coefficients $\beta_{\text{candidate}}$. Check if the target criterion is met for all features in $\mathcal{F}_{\text{target}}$. If not met, expand the candidate set by adding the next $b = n_{\text{min}}/10$ highest-influence examples and repeat. Continue until either the target is achieved or the entire pool is exhausted.

    \item \textbf{Target verification}: After the final iteration, verify that the learned coefficients satisfy the shift criteria. If not achieved, raise an error indicating that the desired shift is infeasible for this dataset with the given constraints and exclude the created train set.
\end{enumerate}

The greedy selection procedure induces distribution shifts across all three canonical types: (i) Covariate shift: Selecting examples based on influence scores changes $p(x)$, as certain regions of feature space are over- or under-represented in the convenience sample. (ii) Label shift: The selection process alters class/target prevalence, changing $p(y)$ relative to the population distribution. (iii) Concept shift: By explicitly manipulating learned coefficients, we directly modify the conditional relationship $p(y|x)$ between features and targets.

\subsubsection{Shift Severity Control}
\label{subsec:knobs}

Our benchmark provides several knobs for controlling the severity of the induced relationship shift:

\begin{itemize}
\item \textbf{Minimum subset size ($\alpha_{min}$).} Smaller values of $\alpha_{min}$ restrict the optimization procedure to a smaller subset of the original training data. This generally produces training samples that are less representative of the underlying population.

\item \textbf{Feature importance.} Perturbing relationships involving highly informative features is expected to have a larger impact on downstream predictive performance than perturbing less informative features. We estimate feature importance using the absolute value of feature--target correlation.

\item \textbf{Coefficient magnitude.} For amplification and attenuation shifts, larger coefficient modifications correspond to stronger departures from the original relationship are therefore expected to result in greater performance degradation.

\end{itemize}

These knobs allow us to evaluate model robustness across a broad spectrum of convenience-sample biases and to analyze how different forms of relationship distortion affect downstream performance. Figure~\ref{fig:knobs} presents an empirical study of the impact of these severity controls on the performance of the evaluated methods.

\section{Experimental Setup}

\subsection{Evaluation}
\label{subsec:eval}

We evaluate seven methods: XGBoost \citep{chen2016xgboost}, LightGBM \citep{ke2017lightgbm}, CatBoost \citep{prokhorenkova2018catboost}, Random Forest \citep{breiman2001random}, Multi-Layer Perceptron (MLP), and TabPFN \citep{hollmann2023tabpfn}, plus Logistic Regression for classification and Ridge Regression for regression. Our baseline suite is deliberately narrower than the one used by
TableShift \citep{gardner2023tableshift}, the closest tabular distribution-shift benchmark to
ours. In Section \ref{app:baselines} we state explicitly which methods we omit and why.

For each dataset, we perform the following evaluation procedure:
\begin{enumerate}
    \item Split data into training (50\%), and test (50\%) sets
    \item Generate convenience training variants using the perturbation strategies described above
    \item For each method:
    \begin{itemize}
        \item Train on in-distribution training set, tune hyperparameters on in-distribution validation set (20\% of the original in-distribution training set), evaluate on test set
        \item Train on convenience training set, tune hyperparameters on convenience validation set (20\% of the original OOD training set), evaluate on test set
        
    \end{itemize}
    \item Compute performance metrics: AUC-ROC for classification (macro-averaged for multiclass), R² for regression
    \item Calculate $\Delta_{perf}$
    \item Remove convenience sets where $\Delta_{perf} < 0.1$ or $(f_\theta^{\text{conv}}, P_{\text{test}}) < -1$ or $(f_\theta^{\text{conv}}, P_{\text{test}}) > 1$.
\end{enumerate}

Furthermore, we remove coefficient-manipulation sets whose induced downstream performance degradation is already subsumed by previously constructed OOD sets. This situation can arise when increasing the magnitude of a feature shift fails to produce additional degradation beyond that observed in an existing OOD set. Because such sets do not expose new model failure modes or provide additional discriminatory power for evaluating robustness, retaining them would introduce redundancy without increasing the benchmark's informational value.

\subsection{Measure of collateral effect}
We measure the collateral effect on non-target coefficients using relative L2 change. Let $\beta_{\text{pop}}$ and $\beta_{\text{conv}}$ denote the coefficient vectors obtained by fitting logistic/ridge regression on the in-distribution and convenience training sets, respectively. For a given shift targeting feature set $\mathcal{T}$, we compute the relative change in non-target coefficients as:

 \begin{equation}
     \text{Relative Change} = \frac{|\beta_{\text{conv}}[\mathcal{T}^c] - \beta_{\text{pop}}[\mathcal{T}^c]|_2}{|\beta_{\text{pop}}[\mathcal{T}^c]|_2}
     \label{eq:collateral}
 \end{equation}
 
 where $\mathcal{T}^c$ denotes the complement of the target feature set. This metric quantifies how much the non-target coefficients change as a fraction of their original magnitude, providing a scale-invariant measure of collateral coefficient drift induced by the greedy selection process. 

\subsection{Baselines}
\label{sec:baselines}

We evaluate seven machine learning methods representing diverse model families, all tuned using the Tree-structured Parzen Estimator (TPE) algorithm~\citep{bergstra2011algorithms} implemented in HyperOpt \citep{bergstra2013hyperopt} with a search budget of 20 evaluations per method-dataset combination. Hyperparameter selection is performed on the convenience training set's 80/20 train-validation split, reflecting the realistic constraint that practitioners have no access to the population distribution during model selection.

\textbf{Logistic Regression / Ridge Regression}: Linear models with regularization. For classification, we use logistic regression; for regression, Ridge regression. Search space: $C \in [0.001, 100]$ (log-uniform) for logistic regression, $\alpha \in [0.001, 1000]$ (log-uniform) for Ridge; penalty type $\in \{\text{L1}, \text{L2}, \text{ElasticNet}, \text{None}\}$ with corresponding L1 ratio $\in [0, 1]$ for logistic regression; solver $\in \{\text{auto}, \text{svd}, \text{cholesky}, \text{lsqr}, \text{sag}, \text{saga}\}$ for Ridge.

\textbf{Random Forest}: Ensemble of decision trees with bootstrap aggregation. Search space: number of trees $\in \{50, 100, 200, 300\}$; max depth $\in \{5, 10, 20, 30, \text{None}\}$; min samples split $\in \{2, 5, 10\}$; min samples leaf $\in \{1, 2, 4\}$; max features $\in \{\text{sqrt}, \text{log2}, \text{None}\}$.

\textbf{XGBoost}: Gradient boosted decision trees with advanced regularization. Search space: number of trees $\in \{50, 100, 200, 300\}$; max depth $\in \{3, 5, 7, 10\}$; learning rate $\in [0.01, 0.3]$ (log-uniform); subsample ratio $\in [0.6, 1.0]$; column subsample ratio $\in [0.6, 1.0]$; min child weight $\in \{1, 3, 5\}$; gamma $\in [0, 0.2]$.

\textbf{LightGBM}: Histogram-based gradient boosting with leaf-wise tree growth. Search space: number of trees $\in \{50, 100, 200, 300\}$; num leaves $\in \{15, 31, 63, 127\}$; learning rate $\in [0.01, 0.3]$ (log-uniform); min child samples $\in \{10, 20, 30\}$; subsample ratio $\in [0.6, 1.0]$; column subsample ratio $\in [0.6, 1.0]$; L1 regularization $\in [0, 1]$; L2 regularization $\in [0, 1]$.

\textbf{CatBoost}: Gradient boosting with ordered target encoding and symmetric trees. Search space: iterations $\in \{50, 100, 200, 300\}$; depth $\in \{4, 6, 8, 10\}$; learning rate $\in [0.01, 0.3]$ (log-uniform); L2 leaf regularization $\in \{1, 3, 5, 7, 9\}$; border count $\in \{32, 64, 128\}$.

\textbf{MLP}: Multi-layer perceptron with ReLU/tanh activations and standard feature scaling. Search space: hidden layer sizes $\in \{(32), (64), (128), (64, 32), (128, 64)\}$; activation $\in \{\text{ReLU}, \text{tanh}\}$; L2 penalty $\in [0.0001, 0.1]$ (log-uniform); learning rate schedule $\in \{\text{constant}, \text{adaptive}\}$; initial learning rate $\in [0.001, 0.1]$ (log-uniform). Maximum 500 iterations with early stopping.

\textbf{TabPFN}: Transformer-based method using in-context learning. No hyperparameter tuning—uses default configuration with 4 ensemble members..

All methods are retrained on the combined train+validation split after hyperparameter selection, then evaluated on the held-out population test set.


\subsection{LLM-Based Coefficient Anomaly Detection}

To assess whether large language models can identify unreasonable feature coefficients, we conducted an experiment where Claude-Haiku-4.5 was presented with OOD coefficients and asked to identify which features seemed most suspicious based on domain knowledge alone. We then calculated the actual coefficient changes ($|IID - OOD|$) for all features and ranked them by magnitude of change. We evaluated whether the LLM's suspicion rankings overlapped with the true rankings of coefficient change, measuring Recall@$K$ as the proportion of the true top-$K$ most-changed features that appeared in the LLM's top-$K$ suspicious features.

\section{Results}
\label{sec:results}

\begin{figure}
    \centering
    \includegraphics[width=1\linewidth]{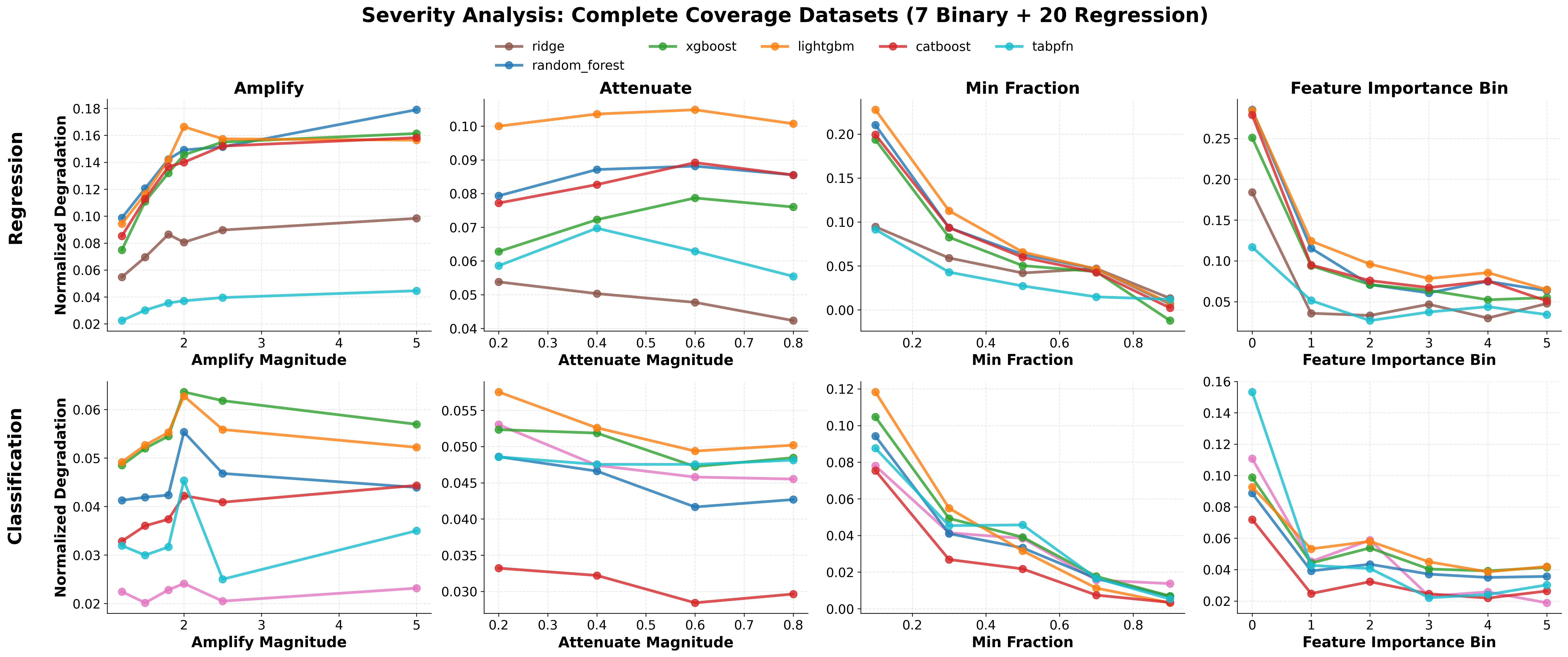}
    \caption{Performance degradation as a function of each shift-severity control. For a given severity knob, we report the average performance of each baseline across all combinations of the remaining severity controls while holding the selected knob fixed. This marginalization isolates the effect of each severity control on downstream performance. MLP and logistic regression performances are removed because of catastrophic failures that hid the performances of other models. We calculate normalized degradation as $\frac{Perf_{IID} - Perf_{OOD}}{Perf_{IID}}$.Detailed performance numbers of all models are available in \S\ref{app:table}}
    \label{fig:knobs}
\end{figure}

\begin{figure}
    \centering
    \includegraphics[width=0.95\linewidth]{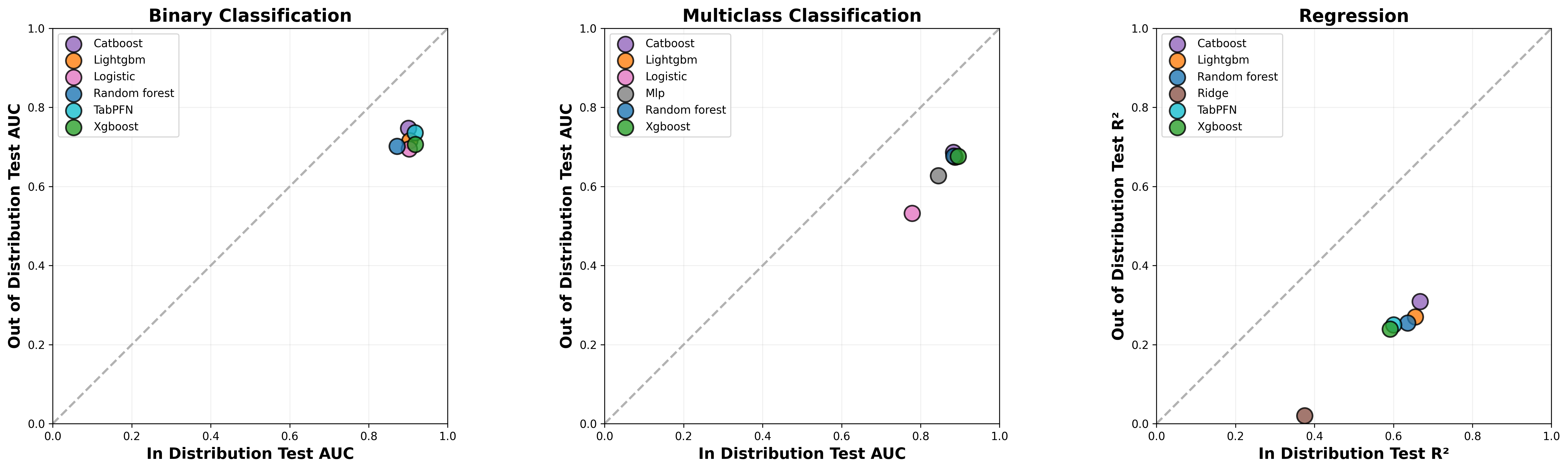}
    \caption{In-distribution versus convenience-sample performance across binary classification, multiclass classification, and regression datasets. Each point represents a baseline. The scatter plots reveal substantial performance degradation under convenience-sample bias relative to the in-distribution setting, highlighting the challenges posed by distorted feature--target relationships.}
    \label{fig:iidvsood}
\end{figure}

\begin{figure}
    \centering
    \includegraphics[width=1\linewidth]{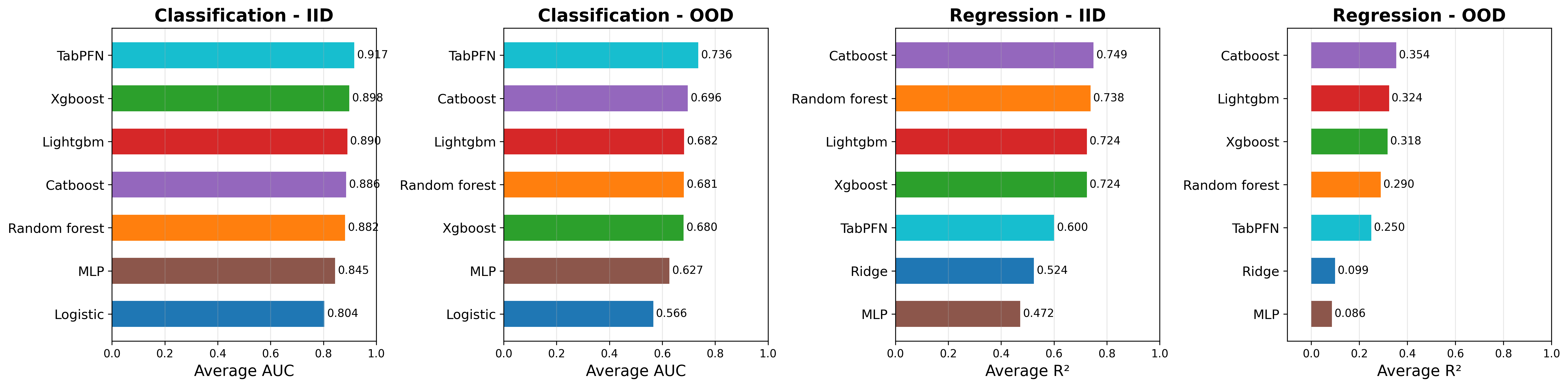}
    \caption{Reported values correspond to the average performance of each method over all the shifts, generated from the classification and regression datasets in the benchmark.}
    \label{fig:placeholder}
\end{figure}

We use \name\ to answer three questions. First, does convenience-sample bias actually
break standard tabular learners, and by how much (\S\ref{sec:agg})? Second, is the induced
shift a realistic, structure-wide shift rather than a single-coefficient artifact
(\S\ref{sec:collateral})? Third, do the severity knobs of \S\ref{subsec:knobs} behave
predictably enough to serve as a difficulty axis (\S\ref{sec:knobs})? Throughout, we report
performance degradation $\Delta_{\text{perf}}$ as defined in Eq.~\ref{eq:shift_gap}.

\subsection{Convenience samples degrade every baseline} \label{sec:agg} Figure~\ref{fig:iidvsood} plots in-distribution against out-of-distribution test performance for every method and task type, averaged over all retained shifts (\S\ref{subsec:eval}). Every baseline sits well below the diagonal, and the methods generally cluster. Averaged over 6,489 classification shifts and 8,970 regression shifts (Figure~\ref{fig:placeholder}), convenience-sample training costs $0.18$--$0.24$ AUC on classification, a $20$--$29\%$ relative drop, and $0.35$--$0.45$ $R^2$ on regression, a $53$--$82\%$ relative drop.

 Figure~\ref{fig:placeholder} breaks this down by method. The plot shows that  the in-distribution ranking does not always survive the shift and lower in-distribution performance does not translate to robustness. Ridge regression and MLP have the weakest in-distribution performance ($R^2 = 0.524$ and $0.472$ respectively) yet suffer among the largest degradations ($\Delta = 0.425$ and $0.386$), falling to under 10\% $R^2$. While a heavily regularized linear model has less capacity to fit complex relationships, when the training data systematically distorts a feature coefficient, Ridge still learns the distortion and fails on population data. This is the central empirical claim of \name. When a training sample reverses or erases a feature--target relationship, nothing in the sample identifies the error. The convenience data are internally consistent, the validation split is drawn from the same biased sample, and no unlabeled population data are available. Correction must therefore come from outside the data which is precisely the setting in which a model with pretrained knowledge can be useful.


\subsection{Collateral effects confirm the shifts are structural}
\label{sec:collateral}


Following Equation \ref{eq:collateral}, we find median relative changes of $26\%$ (amplify), $61\%$
(attenuate), and $55\%$ (flip-sign) on regression tasks, and $109\%$, $107\%$, and $104\%$
respectively on classification tasks. The perturbation propagates through the correlation structure of the feature space rather than remaining local to the targeted feature.


\subsection{The severity knobs control difficulty}
\label{sec:knobs}

Figure~\ref{fig:knobs} shows the effect of varying each severity knob while marginalizing over the others. Since not all datasets produce feasible shifts at every severity level, averaging across all available shifts would result in different dataset subsets being represented at different settings, potentially biasing the comparison. Therefore, we chose seven binary classification, and 20 regression datasets that produced shifts across all shift settings for this analysis. A severity knob is useful for the benchmark if performance degradation changes monotonically with the knob value and if the resulting sweep covers a broad range of difficulty levels. We find that two of the three knobs satisfy both criteria.

\paragraph{Minimum fraction.} $\alpha_{\min}$ controls how small the initial high-influence candidate set is, and therefore how far the selected subset can drift from the population before the target criterion is met. At $\alpha_{\min}=0.1$ degradation reaches $0.09$--$0.23$ $R^2$ for regression and $0.08$--$0.12$ AUC for classification, and it decreases monotonically for every method as the fraction grows which validate $\alpha_{min}$ as a severity control.

\paragraph{Feature importance.} The fourth column of Figure~\ref{fig:knobs} sweeps the feature-importance bin of the targeted feature, where bin0 contains the most important features (by correlation with the target) and bin5 the least. The trend is strongly decreasing: perturbing a top-bin feature costs $0.18$--$0.29$ $R^2$ in regression and $0.07$--$0.11$ AUC in classification, while perturbing a bottom-bin feature costs less than $0.07$ in regression and $0.04$ in classification. This is the expected behavior and validates the bin as a severity control. 

\paragraph{Amplification and attenuation magnitude.} Trends are less consistent. For regression, degradation generally increases with amplification magnitude, but does not consistently decrease as attenuation approaches one and even increases in some cases. For classification, amplification shows no consistent trend, while degradation decreases more consistently as attenuation approaches one. Thus, this severity knob does not reliably control shift difficulty. This inconsistency likely arises from collateral effects: modifying one coefficient can alter the effective influence of other features, making the resulting shift severity unpredictable.


\subsection{LLM-Based Coefficient Anomaly Detection}
Across 300 randomly selected shifts (100 per operation type), the LLM achieved modest performance, with an overall Recall@5 of 37.4\%. Performance varied considerably across operation types. The model was most effective at identifying amplified coefficients (Recall@5 = 42.7\%), followed by sign flips (36.7\%) and attenuations (32.7\%). Compared to a random baseline of 16.8\%, the LLM performs approximately 2.2$\times$ better than chance, demonstrating that LLMs encode domain knowledge that enables them to assess the plausibility of feature--outcome relationships and detect coefficient perturbations that deviate from expected patterns. This capability highlights the potential of LLMs to identify and mitigate biases introduced by convenience sampling.

\section{Conclusion}
Tabular dataset samples are gathered because of cheapness and convenience and rarely represent the greater population. We propose a benchmark for testing model robustness to convenience sampling where the train set is a sample that is non-representative of the general population and the test set represents the general population. We also provide a systematic way to control the severity of the shift to assess model robustness. Across 200 datasets and $\sim 15{,}500$ retained shift variants, seven standard baselines lose between 20\% and 45\% of their performance on average purely as a result of how the training sample was drawn, with the deployment distribution held fixed. Since the biased sample contains no signal that identifies the bias, closing this gap requires information from outside the training data, which is the capability \name\ is built to measure.


\subsection*{AI use statement}

In this work, we used generative AI tools as part of our methodology: Claude-Haiku-4.5 was used in the LLM-based coefficient anomaly detection experiment (\S\ref{sec:results}). We also used Claude-Sonnet-4.5 to assist with portions of the code development process and to generate tables included in the paper.
Additionally, we used ChatGPT-5 to assist with editing and refining the manuscript text. We have reviewed all AI-assisted work: all AI-generated code was reviewed, validated, and, where necessary, modified by the authors. We take responsibility for the final content of this work, including text, claims or artifacts produced with the aid of generative AI, and for the accuracy and validity of all code and analyses.

We did not use AI to Generate synthetic data sets, help develop theoretical models or conceptual frameworks, formulate mathematical claims, provide critical ingredients for proving mathematical claims, assist in the writing of proofs, propose or refine hypotheses, design or provide feedback on research  methodology or experiments, implement methods, assist with translation, clean and reformat dataset, support qualitative and thematic data analysis, or interpret results.


\subsection*{Reproducibility statement}
We provide a command-line interface that enables users to generate shifts by specifying the minimum sample threshold ($\alpha_{min}$), feature-importance bin, coefficient operation, and operation magnitude (for multiplication and attenuation). All preprocessing steps are applied automatically during shift creation. Users can also evaluate any of our baseline methods on a selected shift by providing the paths to the corresponding training and test sets. In addition, the interface supports downloading the complete collection of $ \sim  15{,}500$ shifts.

Using this interface, users can reproduce the results presented in Table~\ref{tab:performance_by_config} and evaluate their own models under shifts of varying types and difficulty levels. A comprehensive usage guide and documentation is included as a README file.

We provide a \href{https://anonymous.4open.science/r/INCONVINIENT-FD51/README.md}{link to the anonymized repository with datasets and full results} in addition to our supplementary material which only includes the scripts.

\bibliography{iclr2027_conference}
\bibliographystyle{iclr2027_conference}

\appendix
\section{Appendix}

\subsection{Inapplicable Methods}
\label{app:baselines}

As mentioned in \S\ref{subsec:eval}, our baseline suite is narrower than TableShift. Most of
the robustness and domain-generalization literature assumes access to information that the
convenience-sampling setting withholds by construction, and identifying which assumptions
fail is itself informative about why this setting is difficult.

\paragraph{Methods requiring multiple training domains.}
Domain generalization methods -- DANN \citep{ganin2016dann}, IRM \citep{arjovsky2019irm},
domain MixUp \citep{zhang2018mixup}, VREx \citep{krueger2021vrex}, DeepCORAL
\citep{sun2016deepcoral}, and MMD \citep{li2018mmd} -- and group DRO over domains
\citep{sagawa2020dro} all require domain labels at training time and at least two training
domains from which an invariance can be inferred. \name\ provides a single convenience
sample with no domain structure, which is the defining feature of the setting we model: a
practitioner with one dataset from one hospital, one platform, or one recruitment pipeline.

The obstacle is more than a missing input. Suppose one partitions the convenience sample
into synthetic environments and runs IRM or VREx on the result. The bias induced by our
greedy selection is global -- the subset is chosen to satisfy a single coefficient criterion
over the whole sample -- so the distorted feature--target relationship is stable across any
such partition. An invariance-seeking objective would identify the manipulated relationship
as invariant and preserve it, which is the opposite of the desired behavior. Invariance is
only informative when some environment exhibits the population relationship; here, none
does.

\paragraph{Methods requiring information about the target distribution.}
Label-shift methods, including group DRO over class labels and adversarial label robustness,
target changes in the marginal $P(y)$ and require either target-domain label proportions or
unlabeled target data to estimate them \citep[e.g.][]{lipton2018bbse}. Our protocol withholds
both. Label shift does occur in \name\ as a by-product of influence-based
selection, but it is a side effect rather than the mechanism, and a method that corrects
$P(y)$ alone leaves the manipulated conditional $P(y \mid x)$ untouched.

\paragraph{Distributionally robust optimization.}
DRO with $\chi^2$ or CVaR geometry \citep{levy2020dro} is the one omitted family that needs
no side information, and it is the omission most worth justifying. These methods minimize
worst-case risk over distributions within a bounded divergence ball of the empirical training
distribution, which amounts to a reweighting of the observed sample. This is exactly the
correction class that the sample-selection literature shows to be insufficient here: when
selection depends only on $X$, importance weighting recovers the population risk
\citep{zadrozny2004learning,cortes2008sample}, but when selection depends on $Y$ or on the $X$--$Y$
relationship -- as our residual-conditioned selection does by design -- the population
conditional is not expressible as a reweighting of the biased sample and is not identifiable
from it. No choice of weights over convenience examples restores a relationship that the
selection removed.

A second, practical obstacle compounds the first. DRO's uncertainty radius is the
hyperparameter that governs how much robustness is purchased at what in-distribution cost,
and our protocol permits tuning only on a validation split drawn from the convenience sample
(\S\ref{sec:baselines}). That split is subject to the same bias as the training data, so it
systematically prefers the smallest radius, collapsing DRO toward ERM. We therefore regard
DRO as a natural addition to a future version of \name\ -- one that would demonstrate this
failure empirically rather than by argument -- but not as a baseline that could plausibly
close the gap reported in \S\ref{sec:agg}.

\paragraph{Tabular neural architectures.}
SAINT \citep{somepalli2022saint}, TabTransformer \citep{huang2020tabtransformer}, NODE
\citep{popov2020node}, and FT-Transformer and tabular ResNet \citep{gorishniy2021revisiting} are
applicable in principle and excluded on cost and expected informativeness rather than
principle. \name\ spans 200 datasets and $\sim 15{,}500$ retained shift variants, each
requiring an IID and an OOD training run with a 20-trial hyperparameter search; the marginal
value of a fifth and sixth deep architecture does not justify that budget, particularly given
that gradient-boosted trees remain the stronger family at this data scale
\citep{grinsztajn2022tree}. More importantly, none of these architectures has access to
information that our included baselines lack. They observe the same anonymous feature matrix
and, like every method in our suite, have no way to know what a column means. Since the only
route back to the population relationship is knowledge external to the biased sample, we
expect architectural variation within the data-driven family to shift the gap marginally
rather than close it. The included MLP and TabPFN bracket this family: a standard deep
network trained from scratch, and a pretrained in-context learner.

\paragraph{Implication.}
Taken together, nearly the entire robustness toolkit is either inapplicable or provably
insufficient in the convenience-sampling regime, because each method assumes an information
source -- multiple environments, group labels, target-domain samples, or a correctable
selection mechanism -- that this setting removes. This is the gap \name\ is designed to
expose, and it motivates evaluating methods whose robustness comes from pretrained knowledge
of what the features mean rather than from the structure of the training sample.

\subsection{Datasets}

The following names all the datasets used in INCONVENIENT.
\subsubsection{Regression (80 datasets)}

\textbf{OpenML (72 datasets):}
\textit{1000 Cameras Dataset}, \textit{3d Estimation Using Rssi of Wlan Dataset}, \textit{3d Estimation Using Rssi of Wlan Dataset Complete 1 Target}, \textit{Abalone}, \textit{Ailerons}, \textit{Airfoil Self Noise}, \textit{Analcatdata Supreme}, \textit{Another Dataset on Used Fiat 500 (1538 Rows)}, \textit{Auction Verification}, \textit{Avocado Sales}, \textit{Bng(echomonths)}, \textit{Bng(lowbwt)}, \textit{Bng(stock)}, \textit{Boston}, \textit{Brazilian Houses}, \textit{Brazilian Houses Reproduced}, \textit{California Housing}, \textit{Cars}, \textit{Concrete Compressive Strength}, \textit{Cookbookreviews}, \textit{Cpmp 2015 Regression}, \textit{Cps1988}, \textit{Cps88wages}, \textit{Cpu Activity}, \textit{Dataset Sales}, \textit{Debutanizer}, \textit{Delta Elevators}, \textit{Diamonds}, \textit{Elevators}, \textit{Energy Efficiency}, \textit{Fifa}, \textit{Forest Fires}, \textit{Goodreads Computer Books}, \textit{Grid Stability}, \textit{Health Insurance}, \textit{House 16h Reg}, \textit{House 8l}, \textit{House Prices Nominal}, \textit{House Sales}, \textit{House Sales Reduced}, \textit{Houses}, \textit{Ieee80211aa Gats}, \textit{Job Profitability}, \textit{Kin8nm}, \textit{Kings County}, \textit{Miami Housing}, \textit{Miamihousing2016}, \textit{Mip 2016 Regression}, \textit{Moneyball}, \textit{Naval Propulsion Plant}, \textit{Onlinenewspopularity}, \textit{Pumadyn32nh}, \textit{Qsar Fish Toxicity}, \textit{Qsar Tid 10980}, \textit{Qsar Tid 11}, \textit{Red Wine}, \textit{Santander Transaction Value}, \textit{Sat11 Hand Runtime Regression}, \textit{Sensory}, \textit{Socmob}, \textit{Solar Flare}, \textit{Space Ga}, \textit{Student Performance Por}, \textit{Sulfur}, \textit{Superconductivity}, \textit{Tecator}, \textit{Treasury}, \textit{Us Crime}, \textit{Video Transcoding}, \textit{White Wine}, \textit{Wind}, \textit{Wine Quality}.

\textbf{Kaggle (1 datasets):}
\textit{Bike Sharing Demand Challange}.

\textbf{TALENT (7 datasets):}
\textit{Bank32nh}, \textit{Cpu Act}, \textit{Cpu Small}, \textit{Pol Reg}, \textit{Puma32h}, \textit{Puma8nh}.

\subsubsection{Binary Classification (39 datasets)}

\textbf{OpenML (23 datasets):}
\textit{Adityakadiwal Water Potability}, \textit{Akshaydattatraykhare Diabetes Dataset}, \textit{Amirhosseinmirzaie Pistachio Types Detection}, \textit{Breast W}, \textit{Climate Model Simulation Crashes}, \textit{Compas Two Years}, \textit{Credit Approval}, \textit{Deependraverma13 Diabetes Healthcare Comprehensive Dataset}, \textit{Diabetes}, \textit{Dresses Sales}, \textit{Eishkaran Heart Disease}, \textit{Fedesoriano Stroke Prediction Dataset}, \textit{Gabrielsantello Cars Purchase Decision Dataset}, \textit{Hashemi221022 Diabetes}, \textit{Kabure German Credit Data With Risk}, \textit{Kandij Diabetes Dataset}, \textit{Saurabh00007 Diabetescsv}, \textit{Shubhamgupta012 Titanic Dataset}, \textit{Tejashvi14 Employee Future Prediction}, \textit{Tejashvi14 Engineering Placements Prediction}, \textit{Tic Tac Toe}, \textit{Uciml Indian Liver Patient Records}, \textit{Uciml Pima Indians Diabetes Database}, \textit{Whenamancodes Predict Diabities}, \textit{Wilt}.

\textbf{Kaggle (11 datasets):}
\textit{Ashishkumarjayswal Diabetes Dataset}, \textit{Barun2104 Telecom Churn}, \textit{Bhavkaur Simplified Titanic Dataset}, \textit{Chirin Africa Economic Banking and Systemic Crisis Data}, \textit{Dileep070 Heart Disease Prediction Using Logistic Regression}, \textit{Jillanisofttech Brain Stroke Dataset}, \textit{Mathchi Diabetes Data Set}, \textit{Nancyalaswad90 Review}, \textit{Vstacknocopyright Blood Transfusion Service Center Data}.

\textbf{TALENT (4 datasets):}
\textit{Alirezachahardoli Bank Personal Loan 1}, \textit{Mahnazarjmand Bank Personal Loan}, \textit{Maryammanoochehry Bank Personal Loan}, \textit{Sanjanchaudhari Bankloan}.

\subsubsection{Multiclass Classification (81 datasets)}

\textbf{OpenML (78 datasets):}
\textit{Aakashjoshi123 Exercise and Fitness Metrics Dataset}, \textit{Abalone}, \textit{Abcsds Pokemon}, \textit{Agirlcoding All Space Missions From 1957}, \textit{Airfoil Self Noise}, \textit{Alexisbcook Pakistan Intellectual Capital}, \textit{Altruistdelhite04 Gold Price Data}, \textit{Amirhosseinmirzaie Countries Life Expectancy}, \textit{Analcatdata Dmft}, \textit{Analcatdata Supreme}, \textit{Ananthr1 Weather Prediction}, \textit{Andrewmvd Udemy Courses}, \textit{Arashnic Time Series Forecasting With Yahoo Stock Price}, \textit{Arnabchaki Data Science Salaries 2023}, \textit{Arnabchaki Indian Restaurants 2023}, \textit{Arnavsmayan Netflix Userbase Dataset}, \textit{Arslanr369 Bitcoin Price 2014 2023}, \textit{Arslanr369 Roblox Stock Pricing 2021 2023}, \textit{Auction Verification}, \textit{Awaiskaggler Insurance Csv}, \textit{Balance Scale}, \textit{Bhanupratapbiswas Bollywood Actress Name and Movie List}, \textit{Bhavkaur Hotel Guests Dataset}, \textit{Car}, \textit{Cars}, \textit{Cmc}, \textit{Concrete Compressive Strength}, \textit{Cpluzshrijayan Milkquality}, \textit{Crxxom Manhwa Dataset}, \textit{Energy Efficiency}, \textit{Forest Fires}, \textit{Gyanprakashkushwaha Laptop Price Prediction Cleaned Dataset}, \textit{Harishkumardatalab Medical Insurance Price Prediction}, \textit{Harshitshankhdhar Imdb Dataset of Top 1000 Movies and Tv Shows}, \textit{Iamsumat Spotify Top 2000s Mega Dataset}, \textit{Iqmansingh Company Employee Dataset}, \textit{Ishadss Productivity Prediction of Garment Employees}, \textit{Jainilcoder Netflix Stock Price Prediction}, \textit{Kanths028 Usa Housing}, \textit{Kin8nm}, \textit{Kingabzpro Cosmetics Datasets}, \textit{Mayankpatel14 Second Hand Used Cars Data Set Linear Regression}, \textit{Mayuriawati Bangalore Chain Restaurants Ratings and Reviews}, \textit{Mfeat Morphological}, \textit{Mirichoi0218 Insurance}, \textit{Moneyball}, \textit{Nikhil1e9 Netflix Stock Price}, \textit{Noordeen Insurance Premium Prediction}, \textit{Oles04 Top Leagues Player}, \textit{Prkhrawsthi Bitcoin Usd Daily Price With Volume 2015 2023}, \textit{Qsar Fish Toxicity}, \textit{Rajyellow46 Wine Quality}, \textit{Receplyasolu 6k Weather Labeled Spotify Songs}, \textit{Red Wine}, \textit{Redwankarimsony Heart Disease Data}, \textit{Rkiattisak Student Performance in Mathematics}, \textit{Rpaguirre Tesla Stock Price}, \textit{Ruchi798 Tv Shows on Netflix Prime Video Hulu and Disney}, \textit{Sabasaeed1953 Stock Prices of 2023}, \textit{Saloni1712 Chatgpt App Reviews}, \textit{Sanjanchaudhari Netflix Dataset}, \textit{Saunakghosh Nba Players Dataset}, \textit{Sbhatti Financial Sentiment Analysis}, \textit{Segment}, \textit{Shreyanshverma27 Imdb Horror Chilling Movie Dataset}, \textit{Shreyapurohit Anime Data}, \textit{Socmob}, \textit{Space Ga}, \textit{Sudarshan6561 Ipl 2023}, \textit{Teertha Ushealthinsurancedataset}, \textit{Timoboz Tesla Stock Data From 2010 to 2020}, \textit{Uciml Red Wine Quality Cortez Et Al 2009}, \textit{Varpit94 Tesla Stock Data Updated Till 28jun2021}, \textit{Vedavyasv Usa Housing}, \textit{Vehicle}, \textit{Vijayvvenkitesh Microsoft Stock Time Series Analysis}, \textit{Visualizing Soil}, \textit{Whenamancodes Students Performance in Exams}, \textit{White Wine}, \textit{Wine Quality}, \textit{Yasserh Wine Quality Dataset}.

\textbf{Kaggle (2 datasets):}
\textit{Awaiskaggler Insurance Csv}, \textit{Bhavkaur Hotel Guests Dataset}.

\subsection{Detailed Performance Numbers}
\label{app:table}

{\scriptsize
3
\begin{longtable}{llllrrrrr}
\caption{Mean performance across all shift configurations. Each row represents the average performance across all datasets for a specific configuration of task type, baseline method, operation, feature importance bin, and minimum fraction.}
\label{tab:performance_by_config} \\
\toprule
Task & Method & Operation & Bin & MinFrac & IID & OOD & Degradation & Shifts \\
\midrule
\endfirsthead

\multicolumn{9}{c}{\tablename\ \thetable\ -- Continued from previous page} \\
\toprule
Task & Method & Operation & Bin & MinFrac & IID & OOD & Degradation & Shifts \\
\midrule
\endhead

\midrule
\multicolumn{9}{r}{Continued on next page} \\
\endfoot

\bottomrule
\endlastfoot
Binary & catboost & amplify & 0 & 0.1 & 0.808 & 0.634 & 0.174 & 8 \\
Binary & catboost & amplify & 0 & 0.3 & 0.893 & 0.704 & 0.190 & 5 \\
Binary & catboost & amplify & 1 & 0.1 & 0.834 & 0.697 & 0.138 & 22 \\
Binary & catboost & amplify & 2 & 0.1 & 0.978 & 0.750 & 0.228 & 2 \\
Binary & catboost & amplify & 3 & 0.1 & 0.767 & 0.627 & 0.140 & 15 \\
Binary & catboost & amplify & 4 & 0.1 & 1.000 & 0.879 & 0.121 & 4 \\
Binary & catboost & attenuate & 0 & 0.1 & 0.890 & 0.708 & 0.182 & 80 \\
Binary & catboost & attenuate & 0 & 0.3 & 0.924 & 0.772 & 0.152 & 40 \\
\end{longtable}
}

\end{document}